\documentclass[11pt]{article}
\usepackage[final]{acl}

\usepackage{times}
\usepackage{latexsym}
\usepackage[T1]{fontenc}
\usepackage[utf8]{inputenc}
\usepackage[protrusion=true,expansion=false]{microtype}
\usepackage{graphicx}
\usepackage{amsmath, amssymb}
\usepackage{booktabs}

\usepackage{xspace}
\usepackage[table]{xcolor}
\usepackage{bbm}
\usepackage[capitalize,noabbrev]{cleveref}
\usepackage{adjustbox}
\usepackage{etoolbox}

\usepackage{array}

\newcolumntype{L}[2]{%
    >{\adjustbox{angle=#1,raise=\depth,lap=(#2)-\width}\bgroup}%
    r%
    <{\egroup}%
}
\newcommand*\rot{\multicolumn{1}{L{-45}{2.5em}}}

\newcommand{\game}[2][]{%
  \ifstrempty{#1}{\texttt{#2}}{\texttt{#2\_{P#1}}}%
}

\newcommand{\imagegame}[1]{\game[#1]{img}}
\newcommand{\matchitascii}[1]{\game[#1]{mia}}
\newcommand{\referencegame}[1]{\game[#1]{ref}}
\newcommand{\taboo}[1]{\game[#1]{tab}}

\definecolor{mshcolor}{HTML}{2E5C8A}

\newcommand{\skg}[1]{\textcolor{purple_taxonomy}{#1}}
\newcommand{\srg}[1]{\textcolor{green_taxonomy}{#1}}

\newcommand{\eg}{\textit{e.g.},\xspace}
\newcommand{\ie}{\textit{i.e.},\xspace}

\crefname{section}{\S}{\S\S}
\crefname{table}{Tab.}{Tab.}
\crefname{figure}{Fig.}{Figs.}
\crefname{algorithm}{Alg.}{}
\crefname{equation}{Eq.}{Eq.}
\crefname{appendix}{App.}{}
\crefname{theorem}{Theorem}{}
\crefname{prop}{Proposition}{}
\crefname{definition}{Def.}{}
\crefname{cor}{Corollary}{}
\crefname{observation}{Observation}{}
\crefname{assumption}{Assumption}{}
\crefname{hyp}{Hyp.}{Hypotheses}
\crefformat{section}{\S#2#1#3}
\crefname{namedtheorem}{Hyp.}{Hypotheses}

\newcommand{\hc}[2]{\cellcolor{cyan!#1}#2}  

\renewcommand{\vec}{\mathbf}

\title{Investigating Knowledge Transfer Across Interactive Dialogue Games}

\author{ 
\textbf{Filippo Momentè}\textsuperscript{$\otimes$}\thanks{Joint first authorship. Correspondence to: \href{mailto:filippo.momente@unitn.it}{filippo.momente@unitn.it}},
\textbf{Mir Nafis Sharear Shopnil}\textsuperscript{$\diamond$}\textsuperscript{$\ast$},
\textbf{Andrea de Varda}\textsuperscript{$\S$},\vspace{0.1em}\\
\textbf{Pavel Merinov}\textsuperscript{$\ddagger$}, 
\textbf{Raffaella Bernardi}\textsuperscript{$\ddagger$},
\textbf{Oswald Lanz}\textsuperscript{$\ddagger$},
\textbf{Alessandro Suglia}\textsuperscript{$\triangle$},\vspace{0.1em}  \\
\textbf{Alessandro Torcinovich}\textsuperscript{$\ddagger$},\\
\textsuperscript{$\otimes$}University of Trento,
\textsuperscript{$\diamond$}Technovative Solutions Ltd,
\textsuperscript{$\S$}Massachusetts Institute of Technology,\\ 
\textsuperscript{$\ddagger$}Free University of Bozen Bolzano,
\textsuperscript{$\triangle$}University of Edinburgh
}
\begin{document}
\maketitle
\begin{abstract}
    Dialogue games represent a challenging setting where complex cognitive skills are required to accomplish tasks while coordinating with other players. Considering that language represents an interface for both understanding the game rules and executing actions, it is reasonable to assume that training on a specific language game will enhance specific capabilities that might be relevant for other tasks as well. Motivated by this rationale, in this paper, we investigate how knowledge transfers across different dialogue games. We study \emph{transferability} by finetuning LLM models on games from the \texttt{clembench} suite~\cite{chalamalasetti-etal-2023-clembench} and performing two analyses: i) we derive a task-transferability graph using a binary integer optimization program from~\citet{zamir2018taskonomy}, using task performance as the main metric; and ii) we compute task vectors~\cite{ilharco2022editing} for each game to study similarities across finetuned models and their task transferability. In our first analysis, we find that some games benefit more from transfer than finetuning, and that the visuospatial family (\eg exploration games) transfers best. With our task vector analysis instead, we find that similarity-based approaches capture game-role relationships but almost no transferability patterns, suggesting that more complex metrics are required. Code available at \url{https://anonymous.4open.science/r/yk6ub2dtwl-E76C} 
\end{abstract}

\section{Introduction}

Dialogue games have posed themselves as complementary to the traditional, reference-based paradigm~\cite[\eg MMLU from][]{Hendrycks2020MeasuringMM} that has long been central in the evaluation of LLMs~\cite{dialoguegames,bertolazzi-etal-2023-chatgpts, momente-etal-2025-triangulating,suglia2024visually}. Their interactive, multi-turn, goal-oriented nature provides for an interesting environment to evaluate agents' capability of understanding language and using it productively to accomplish specific tasks, rather than just retrieving knowledge.

In dialogue games, language expresses both the goals and the constraints (\ie rules) that are required for their completion. The complexity of the game structure requires the exercise of a variety of skills to play and makes its analysis particularly challenging. For instance, understanding the mechanisms employed by a model to reach a game goal implies disentangling reasoning processes from rule-following, identifying the contributions of individual turns towards the objective, and the global strategy applied to solve the task.


Intuitively, we may expect that whenever two games match in any of the aforementioned aspects, training on one game would also help in playing the other. In the literature, there is evidence of such positive knowledge transfer \cite{liu2026spiral, hu2025lmgame, orak}. However, experiments have been performed only on a limited number of tasks, and a systematic study of these effects across a variety of dialogue games is currently missing.


In this work, we are therefore interested in further investigating relationships among dialogue games, and studying their \emph{transferability}, \ie the degree to which specific game knowledge can be used to also play other games. To this aim, we focus on the popular \texttt{clembench} evaluation framework~\cite{chalamalasetti-etal-2023-clembench}, and we adapt the Taskonomy framework~\cite{zamir2018taskonomy}, finetuning LLMs on specific game tasks and obtaining a series of graph representations that capture the underlying structure of knowledge transfer relationships.
%
%
With our approach, we showcase that it is possible to find relevant relationships between different dialogue games. To our surprise, we observe several cases where finetuning on a given game may not be the best way to improve at it, and that it would be advisable to train our model on another game instead. We also report that the family of visuospatial games, involving reasoning on spatial information (\eg exploration games), provides transferable knowledge to purely verbal games (\eg Taboo). Thanks to our analysis, we also identify several situations of asymmetric transfers, where finetuning on a given game $A$ generalizes to another game $B$, but doing the opposite degrades performance. These results support the fact that dialogue games require complex natural language understanding and reasoning skills that are instantiated in different ways by each game.

We also provide a complementary weight-space analysis of transferability and the connection between transferability and weight updates of finetuned models. To do so, we leverage task vectors~\cite{ilharco2022editing} to determine the difference in parameter space between a finetuned model on a specific game and its baseline, \ie the model used as the starting point. By comparing task vectors associated with each game, we find that weight similarity does not explain transfer relationships but rather offers a fingerprint of the game and the player role.


Taken together, the taxonomy and the weight-space analysis offer two complementary views of the same phenomenon: weight similarity records what was trained, while the direction of transfer must be measured by playing the games. Beyond the specific relationships we map, this gives an interpretability handle on what game-specific finetuning does to a model, a step towards understanding the mechanisms LLMs employ while playing dialogue games.




\section{Related Work}

\paragraph{Dialogue Games for LLM Evaluation.}
Dialogue games are well-structured activities with a \emph{goal state} which players have to reach mostly through the production and use of language~\cite{levin1977dialogue,dialoguegames}. They are typically multi-turn and can involve multiple players with different roles. While prior work has demonstrated their value in revealing capabilities that static benchmarks fail to surface, studies have generally treated game performance as an end in itself~\cite{chalamalasetti-etal-2023-clembench, hakimov-etal-2026-price, qiao2023gameeval}. Similar to~\citet{momente-etal-2025-triangulating}, we instead use performance across a structured collection of dialogue games as a lens to study the internal organisation of model capabilities and how they transfer across tasks.

\paragraph{Interactive and Collaborative Games as Assessment Probes.} Game-based approaches have a well-established history in assessing human cognitive and collaborative skills. Prior work has leveraged gamified environments to measure Collaborative Problem Solving (CPS) competencies such as strategy and perspective taking~\cite{stoeffler2020gamified}, and serious games have been used to evaluate soft skills including communication and teamwork~\cite{marengo2025machine}. Such studies derive rich behavioural signals from naturalistic interaction traces without disrupting gameplay. Our work shares this methodological intuition — that interactive, collaborative environments make for uniquely informative probes — but shifts the subject of assessment from human participants to language models, employing dialogue games as structured instruments to study LLM capabilities across a diverse range of tasks.

\paragraph{Transfer Learning Across Tasks.} The broader question of which tasks benefit from shared training has been extensively studied in NLP. \citet{pruksachatkun2020intermediate} and~\citet{vu-etal-2020-exploring} systematically characterised intermediate-task transfer, showing that transfer relationships are often asymmetric and difficult to predict from surface-level task similarity alone. To organise such relationships structurally, we adapt the Taskonomy framework~\cite{zamir2018taskonomy}, originally developed for computer vision, to the dialogue game domain. Our work is the first to apply this kind of structured transfer analysis systematically across a broad suite of dialogue games.

\paragraph{Parameter Space Analysis and Mechanistic Interpretability.} A growing body of work has attempted to characterise what fine-tuning changes inside a model, both from a parameter-level and a mechanistic perspective. Building on the task vector framework of~\citet{ilharco2022editing}, several studies have examined how task-specific adaptations interact and compose in weight space ~\cite{yadav2023ties, wortsman2022model}, while~\citet{aghajanyan2021intrinsic} show that such adaptations tend to occupy low-dimensional subspaces — providing theoretical grounding for interpreting task vector geometry. Complementing this parameter-level view, mechanistic interpretability work has probed the internal structure of transformer models more directly~\cite{elhage-etal-2021-mathematical, dai-etal-2022-knowledge, geva-etal-2021-transformer}. Particularly relevant to our findings is the work of~\citet{prakash-etal-2024-finetuning}, who show that fine-tuning tends to enhance pre-existing internal mechanisms rather than introduce new ones, offering a potential explanation for why some game-specific adaptations generalise more broadly than others. 

\section{Investigating Knowledge Transfer}






\subsection{Dialogue Games}
\paragraph{Terminology.}
Dialogue games are instantiated through prompts, composed of a fixed template and variable parameters (\eg the word to guess, the forbidden ones, etc.). Each of these prompts constitute an \emph{instance} of that game.
Players play a game instance by performing, at each turn one among all the possible actions available at that point, until reaching a \emph{final} or an \emph{aborted state}. The sequence of the actions taken at any turn is collected in an \emph{episode}. In dialogue games, actions are expressed through language, making an episode a multi-turn conversation. Episodes are recorded into transcripts, and used to construct a dataset.

\texttt{clembench}~\cite{chalamalasetti-etal-2023-clembench} provides a dataset of textual and multimodal dialogue games and a game evaluation framework for language models. In two-player games, the players never communicate directly and a game master mediates their communication, by preprocessing messages according to game logic before sending them. Additionally, the players may be shown different prompt templates reflecting their different roles within a game. This affects their context received at each turn, leading them to develop different \emph{perspectives}, or points of view, on the episode that is being played. The framework records transcripts for played episodes, taking into account each player's perspective.

\paragraph{Game Selection and Dataset Preparation.}
First, we collect game transcripts of the $17$ textual dialogue games in the open-source \texttt{clembench-runs} repository~\cite{clembenchruns} (\texttt{1.6}, \texttt{2.0}, \texttt{3.0}) obtained by having multiple models play several instances of the \texttt{clembench} dialogue games. We remove duplicate transcripts, those with more than $3000$ tokens (as measured by the Qwen3.5-4B model's tokenizer \cite{qwen3.5}), those associated with episodes that did not terminate succesfully. In two-player games, transcripts are split into two independent conversations mirroring the players roles (from now on, identified with \texttt{P1} and \texttt{P2}). We obtain a total of $27$ role-specific \emph{tasks}.
To obtain a balanced dataset, we exclude tasks with fewer than $800$ samples. Additionally, we exclude two tasks for their similarities with the other considered games, resulting in a final selection of $15$ role-specific tasks associated with $9$ games.

We finetune our model with a standard autoregressive loss on the transcripts. In particular, we hold out $20$ transcripts per task to create the validation set. We do the same for the test set, keeping only the game instances to test the game play of the trained model. With the remaining transcripts, we build the training set, subsampling each task to the smallest task set size, resulting $585$ samples per task.

\paragraph{Game Overview.}
We group the selected games into two families: \emph{verbal} operating over words; \emph{visuospatial} operating over grids and spatial configurations. Such distinction coarsely characterize the primary cognitive demands of each game, and appears both in models of working memory~\cite{BADDELEY197447}, and in large-scale factor analyses of cognitive abilities~\cite{carroll1993human}. In what follows, we provide a brief description of each game and their selected role-specific tasks.

\texttt{codenames}: verbal, two-player. The players compete together against a programmatically defined team. Both teams are provided with the same list of words, where some are associated with one team, some with the other. Only one team member is aware of the distinction. At each turn, this player generates a clue associated to a subset of words in the team list, helping the other into guess. Found words are removed from the list until one team guesses all of them, or a designed ``killer word'' is guessed.``Distractor words'' are also present.
\begin{itemize}
  \item \texttt{P1}: identifies targets and provides clues.
  \item \texttt{P2}: makes the guesses.
\end{itemize}
\texttt{guesswhat}: verbal, two-player. A player is given a list of words and needs to guess a target word by asking questions to the other player who knows the answer and is only allowed to reply with``yes''/``no''.
\begin{itemize}
  \item \texttt{P1}: asks questions and makes guesses.
  \item \texttt{P2}: answers ``yes'' or ``no''.
\end{itemize}
\texttt{taboo}: verbal, two-player. A player helps the other to guess a word by providing clues at each turn. Clues must not contain the target and other related words.
\begin{itemize}
  \item \texttt{P1}: provides the clues.
  \item \texttt{P2}: makes a guess.
\end{itemize}
\texttt{wordle\_withclue}: verbal, single-player. The player tries to guess a $5$-letter word based on clues and feedback indicating which guessed letters are present and whether they are in the correct position.
\texttt{wordle\_withcritic} (VB): two-player. Identical to \texttt{wordle\_withclue} but a second player provides additional feedback after each attempt.
\begin{itemize}
  \item \texttt{P1}: makes a guess.
  \item \texttt{P2}: provides the feedback.
\end{itemize}

\texttt{adventuregame}: visuospatial, single-player. A player explores and interacts with a simulated environment composed by multiple rooms, takes and replaces objects to reach its goal.
\texttt{imagegame}: visuospatial, two-players, collaborative. One player receives a representation of an ASCII grid as input and has to guide the other player to reconstruct it turn by turn starting from an empty one.
\begin{itemize}
  \item \texttt{P1}: provides the instructions.
  \item \texttt{P2}: draws a the grid based on the received instructions.
\end{itemize}
\texttt{matchit\_ascii}: visuospatial, two-player. Each player is provided with an ASCII grid and has to understand whether it is the same the other player sees or not by asking questions. Both players cover the same role, although the input grid may differ.
\texttt{referencegame}: visuospatial, two-player. One player is provided with an ASCII grid to describe and another one has to exploit the description to pinpoint it among three options.
\begin{itemize}
  \item \texttt{P1}: describes the target grid.
  \item \texttt{P2}: makes a guess.
\end{itemize}

\subsection{Taxonomy Modeling}
We are given a set of tasks $\mathcal{T}$ and a training budget. Our objective is to train some models on a subset of tasks $S \subseteq \mathcal{T}$ constrained by the budget, in order to maximize the overall performance in $\mathcal{T}$. Then, from the solution of such problem, structural information on the transferability between tasks emerges, \ie which task knowledge learned by a model transfers better to other tasks.

More formally, we define a \emph{task-transferability graph} as a weighted directed hypergraph $G = (V, E, w)$ whose vertex set $V = \mathcal{T}$, $|V| = n$ represents tasks we want to analyze. Given the set of edges $E = V \times V$, we define a weighting function $w: V \times V \rightarrow \mathbb{R}_{\ge 0}$ that associates each directed edge $(s, t) \in E$ to a \emph{transfer performance}, \ie the performance obtained on a \emph{target task} $t$ by a model trained on the \emph{source task} $s$.

Given a set $S \subseteq V$ of source tasks, its \textit{transfer set} is defined as:
\begin{equation}
    T = \left\{t \in V \mid \text{max}_{s \in S} w(s, t) > 0 \right\},
\end{equation}
and its \emph{collective transfer performance} is:
\begin{equation}
    P(S) = \sum_{t \in T} \max_{v \in V} w(v, t).
\end{equation}

Given a budget $\gamma \in \mathbb{N}^+$, the \emph{collective transferability problem (CTP)} requires then finding a set $S^\ast \subseteq V$, $|S^\ast|$ maximizing the collective transfer performance, \ie $P(S^\ast) \ge P(S)$, $\forall S \subseteq V$. The problem can be defined with the following binary integer program:
\begin{alignat}{2}
    \max_\vec{x} \quad & \vec{w}^\top\vec{x}  \label{eq:objective} \\
    \text{subject to} \quad & \nonumber \\
    \vec{A}\vec{x} & \preceq \vec{b},\ \vec{x} &\in \{0, 1\}^{|E| + |V|}. \label{eq:constraints}
\end{alignat}
where:
\begin{equation}
    w_k =
    \begin{cases}
      w(s, t) & \text{if }  k \le |E| \land \sigma(k) = s \land \tau(k) = t  \\
      0  & \text{otherwise},
    \end{cases}
\end{equation}
with $\sigma(k) = k \text{ mod } n$ and $\tau(k) = k \text{ div } n$ returning respectively the source and the target tasks of an edge $(\sigma(k), \tau(k))$.

\begin{table*}[!t]
\centering
\setlength{\tabcolsep}{3.5pt}
\renewcommand{\arraystretch}{1.15}
\resizebox{\textwidth}{!}{%
\begin{tabular}{c l ccccccccccccccc}
 & & \multicolumn{15}{c}{\textbf{Target}} \\[0.5ex]
 & 
 & \rot{\srg{\texttt{adventuregame}}}
 & \rot{\skg{\texttt{codenames\_P1}}}
 & \rot{\skg{\texttt{codenames\_P2}}}
 & \rot{\skg{\texttt{guesswhat\_P1}}}
 & \rot{\skg{\texttt{guesswhat\_P2}}}
 & \rot{\srg{\texttt{imagegame\_P1}}}
 & \rot{\srg{\texttt{imagegame\_P2}}}
 & \rot{\srg{\texttt{matchit\_ascii\_P1}}}
 & \rot{\srg{\texttt{referencegame\_P1}}}
 & \rot{\srg{\texttt{referencegame\_P2}}}
 & \rot{\skg{\texttt{taboo\_P1}}}
 & \rot{\skg{\texttt{taboo\_P2}}}
 & \rot{\skg{\texttt{wordle\_withclue}}}
 & \rot{\skg{\texttt{wordle\_withcritic\_P1}}}
 & \rot{\skg{\texttt{wordle\_withcritic\_P2}}} \\
\midrule
\multirow{15}{*}{\rotatebox{90}{\textbf{Source}}}
 & \srg{\texttt{adventuregame}}        & \hc{20}{0.50} & \hc{12}{0.29} & \hc{10}{0.25} & \hc{4}{0.10} & \hc{35}{0.88} & \hc{3}{0.07} & \hc{4}{0.10} & \hc{8}{0.20} & \hc{2}{0.05} &      &      &      &      &      &      \\
 & \skg{\texttt{codenames\_P1}}        &      & \hc{11}{0.27} & \hc{12}{0.30} & \hc{4}{0.10} & \hc{31}{0.77} & \hc{0}{0.01} & \hc{5}{0.12} & \hc{8}{0.20} &      &      &      &      &      &      & \hc{0}{0.01} \\
 & \skg{\texttt{codenames\_P2}}        &      & \hc{4}{0.10}  & \hc{9}{0.22}  &      & \hc{31}{0.78} & \hc{2}{0.04} & \hc{4}{0.09} & \hc{4}{0.10} &      &      & \hc{1}{0.02} &      &      &      &      \\
 & \skg{\texttt{guesswhat\_P1}}        &      &      & \hc{11}{0.27} &      & \hc{33}{0.82} &      & \hc{6}{0.15} &      & \hc{2}{0.05} &      & \hc{2}{0.05} & \hc{1}{0.02} &      & \hc{3}{0.08} &      \\
 & \skg{\texttt{guesswhat\_P2}}        &      & \hc{10}{0.25} & \hc{20}{0.50} &      & \hc{36}{0.90} &      & \hc{1}{0.03} & \hc{2}{0.05} & \hc{4}{0.10} &      & \hc{4}{0.10} &      &      &      &      \\
 & \srg{\texttt{imagegame\_P1}}        &      & \hc{7}{0.17}  & \hc{12}{0.29} &      & \hc{35}{0.88} & \hc{5}{0.12} & \hc{8}{0.19} & \hc{4}{0.10} & \hc{8}{0.20} &      & \hc{3}{0.08} &      &      &      &      \\
 & \srg{\texttt{imagegame\_P2}}        &      & \hc{7}{0.17}  & \hc{14}{0.36} & \hc{19}{0.47} & \hc{34}{0.85} &      & \hc{9}{0.23} & \hc{4}{0.10} & \hc{6}{0.15} &      & \hc{3}{0.07} & \hc{1}{0.02} &      &      &      \\
 & \srg{\texttt{matchit\_ascii\_P1}}   &      & \hc{11}{0.27} & \hc{17}{0.42} & \hc{7}{0.18} & \hc{33}{0.82} & \hc{3}{0.07} & \hc{2}{0.06} & \hc{8}{0.20} &      & \hc{2}{0.05} & \hc{3}{0.07} &      &      &      &      \\
 & \srg{\texttt{referencegame\_P1}}    &      & \hc{10}{0.25} & \hc{12}{0.30} &      & \hc{35}{0.88} &      &      &      & \hc{2}{0.05} & \hc{8}{0.20} & \hc{2}{0.05} & \hc{1}{0.02} & \hc{0}{0.01} &      & \hc{1}{0.03} \\
 & \srg{\texttt{referencegame\_P2}}    &      & \hc{18}{0.44} & \hc{14}{0.36} & \hc{6}{0.14} & \hc{33}{0.83} & \hc{2}{0.04} & \hc{1}{0.02} & \hc{8}{0.20} &      & \hc{6}{0.15} & \hc{3}{0.07} &      &      &      &      \\
 & \skg{\texttt{taboo\_P1}}            &      &      & \hc{13}{0.33} &      & \hc{33}{0.83} &      & \hc{1}{0.02} & \hc{4}{0.10} &      &      & \hc{2}{0.05} &      &      &      &      \\
 & \skg{\texttt{taboo\_P2}}            &      & \hc{7}{0.18}  & \hc{12}{0.29} &      & \hc{31}{0.77} &      & \hc{2}{0.05} & \hc{2}{0.05} & \hc{4}{0.10} &      &      & \hc{2}{0.06} &      &      &      \\
 & \skg{\texttt{wordle\_withclue}}     &      & \hc{3}{0.08}  & \hc{12}{0.31} & \hc{6}{0.14} & \hc{33}{0.82} & \hc{2}{0.06} & \hc{3}{0.08} &      &      &      & \hc{4}{0.10} &      &      &      &      \\
 & \skg{\texttt{wordle\_withcritic\_P1}} &    & \hc{4}{0.09}  & \hc{15}{0.38} & \hc{9}{0.23} & \hc{34}{0.85} & \hc{3}{0.08} & \hc{4}{0.11} &      &      &      & \hc{6}{0.15} &      &      &      &      \\
 & \skg{\texttt{wordle\_withcritic\_P2}} &    & \hc{5}{0.12}  & \hc{13}{0.33} & \hc{9}{0.22} & \hc{34}{0.85} & \hc{2}{0.05} & \hc{4}{0.10} & \hc{6}{0.15} &      &      & \hc{2}{0.04} & \hc{1}{0.02} &      &      &      \\
\bottomrule
\end{tabular}%
}
\caption{Filtered transfer modeling matrix. Tasks are divided into \textcolor{green_taxonomy}{visuospatial}, and \textcolor{purple_taxonomy}{verbal}; \texttt{P1}/\texttt{P2} denotes player roles, while empty cells correspond to specialists underperforming the baseline.}
\label{tab:filtered_quality_matrix}
\end{table*}
We have three constraint types.

\paragraph{Constraint I} If an edge is selected, \ie $x_k = 1$, for any $k \le |E|$, then its corresponding source node must be selected too. $\forall i \le |E|$:
\begin{align}
   a_{i, k} & = 
   \begin{cases}
       1 & \text{if } k = i \\
       -1 & \text{if } k - |E| = \sigma(i) \\
       0 & \text{otherwise},
   \end{cases} \\
   b_i & = 0.
\end{align}

\paragraph{Constraint II} Each target task has at most one in-transfer (\ie at most one incoming source task). $\forall j \le |V|$:
\begin{align}
   a_{|E| + j, k} & =
   \begin{cases}
      1 & \text{if } k \le |E| \land \tau(k) = j\\
      0 & \text{otherwise},
   \end{cases}
   \\
   b_{|E| + j} & = 1.
\end{align}

\paragraph{Constraint III} The source set must contain at most $\gamma$ source tasks.
\begin{align}
   a_{|E| + |V| + 1, k} & = 
   \begin{cases}
     1 & \text{if } k > |E|\\
     0 & \text{otherwise},
   \end{cases}
   \\
   b_{|E| + |V| + 1} & = \gamma.
\end{align}

The optimum $\vec{x}^\ast$ entails then the searched graph structure.

\paragraph{Transferability Scores.} We now discuss how to obtain the transferability scores required by our CTP. We finetune instances of the same baseline model on each of the tasks in our dataset, obtaining \emph{task specialists}. Each specialist is then evaluated on all the tasks, computing its mean \emph{quality score ($qs$)}, a metric defined  in the \texttt{clembench} paper to disentangle game-playing competence from rule-following.

The baseline model scores are then subtracted from each other score, to obtain an estimate of the improvement attributable by the training. Some specialists perform worse than the baseline on certain games, thus, we remove the corresponding entries of the matrix, obtaining a filtered \emph{transfer modeling matrix} shown in \cref{tab:filtered_quality_matrix}. 

\subsection{Weight Space Analysis} \label{sec:weightspace_method}
We complement the previous analysis by taking a look at the differences induced by model finetuning on a given task.

For each task and specialist, we compute its task vector~\cite{ilharco2022editing} $\tau_t = \theta_t - \theta_{\textit{Base}}$, with $\theta_t$ and $\theta_{\textit{Base}}$ being respectively the parameters of the specialist and the base model. $\tau_t$ then represents the change induced by the training.\footnote{We perform finetuning over the entire base LLM (no adapters), so $\tau_t$ is defined over the entire network weights.} 
\begin{figure*}[t]
    \centering
    \begin{minipage}{0.5\textwidth}
        \centering
        \includegraphics[width=\linewidth]{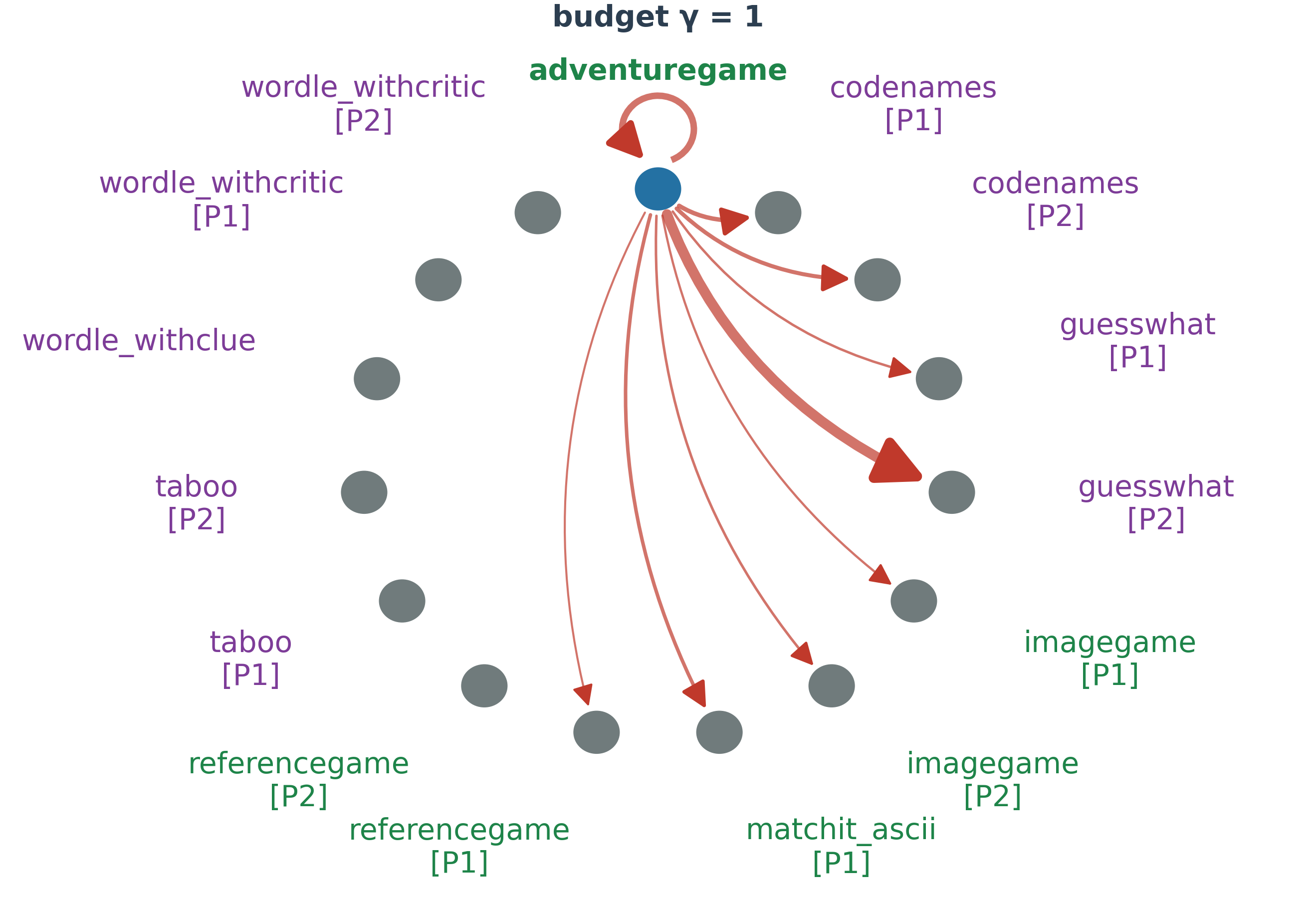}
    \end{minipage}%
    \begin{minipage}{0.5\textwidth}
        \centering
        \includegraphics[width=\linewidth]{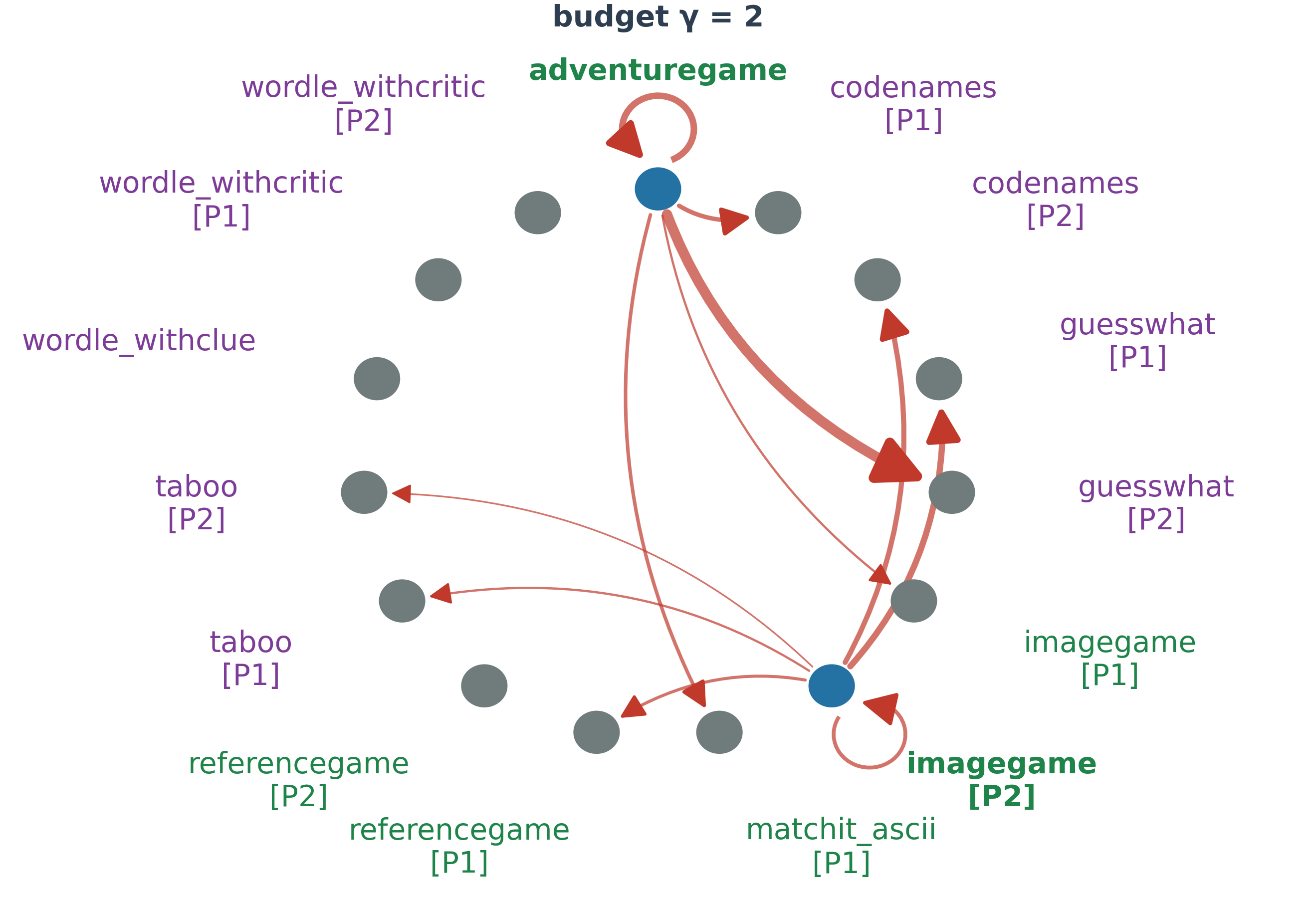}
    \end{minipage}

    \vspace{1em}

    \begin{minipage}{0.5\textwidth}
        \centering
        \includegraphics[width=\linewidth]{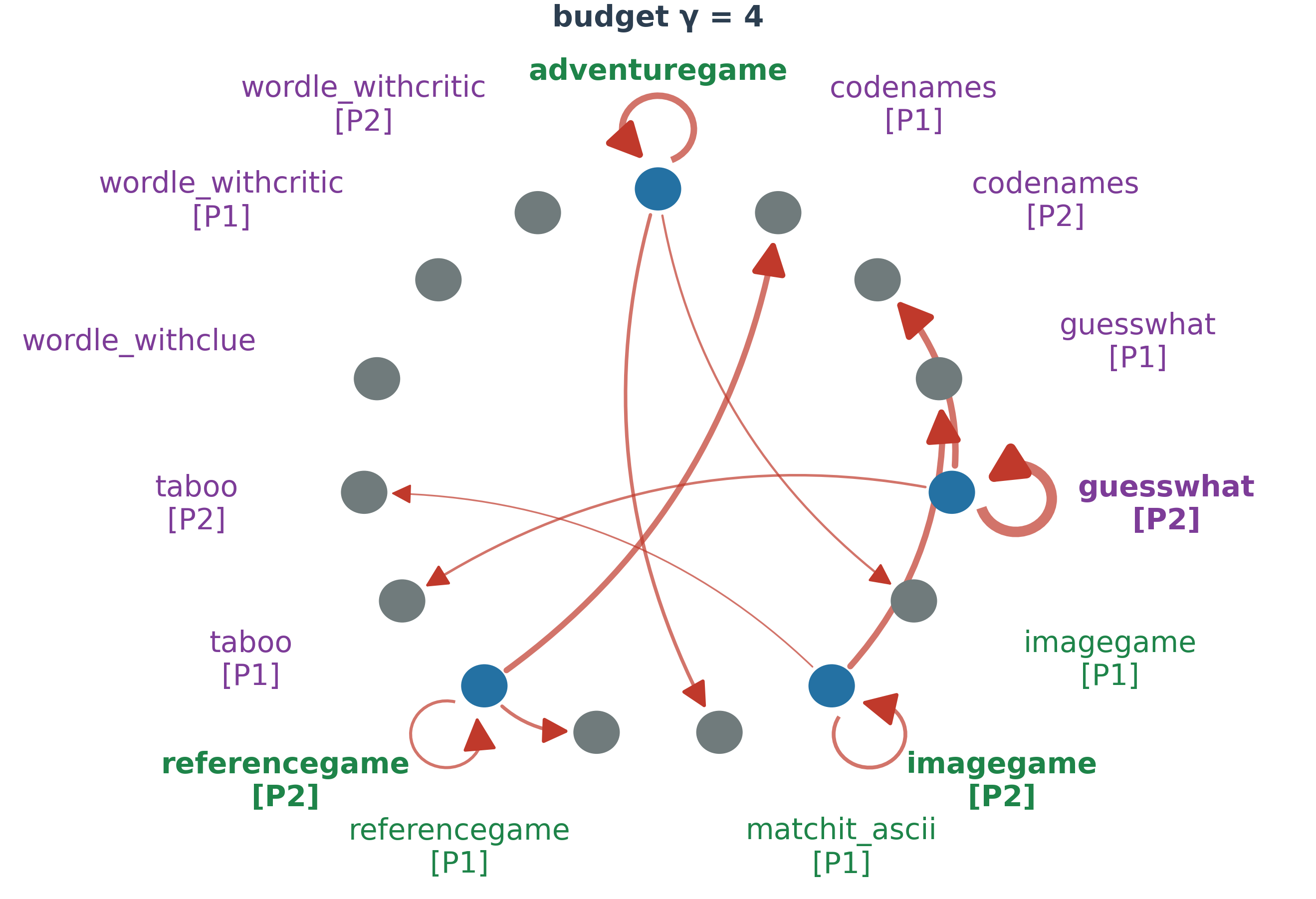}
    \end{minipage}%
    \begin{minipage}{0.5\textwidth}
        \centering
        \includegraphics[width=\linewidth]{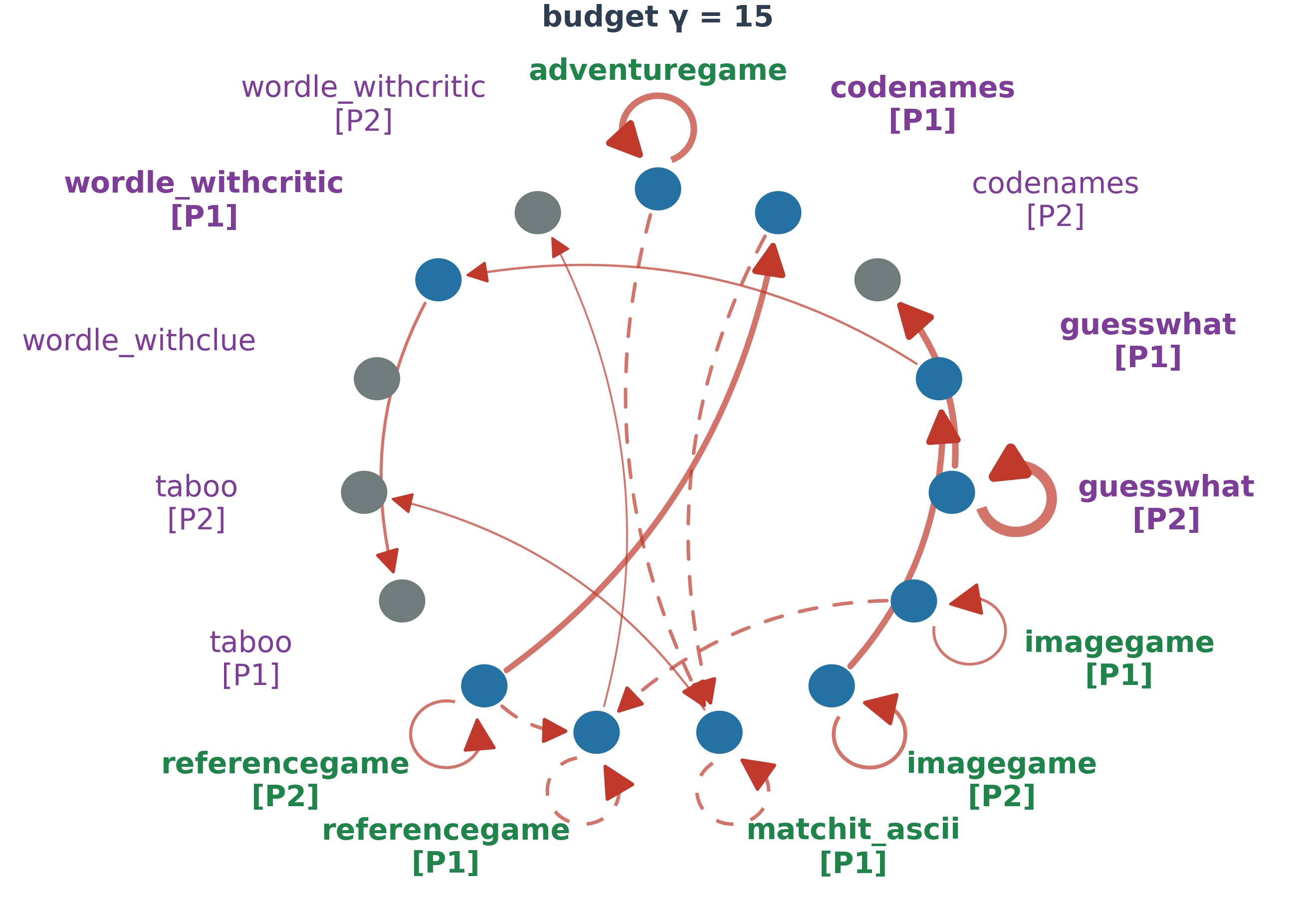}
    \end{minipage}

    \caption{Taxonomies across different $\gamma$ values. Tasks are divided into \textcolor{green_taxonomy}{visuospatial}, and \textcolor{purple_taxonomy}{verbal}; \textcolor{blue_taxonomy}{blue} encodes the transfer set. \textcolor{red_taxonomy}{Edges} represent transfers, with edge thickness proportional to the improvement. Dashed edged indicate multiple optimal transfers for the target. Edges depart prevalently from vertices of visuospatial tasks, indicating their versatile transferability across all of the considered games.}
    \label{fig:qs_taxonomies}
\end{figure*}

We then compare the task vectors to identify similar patterns. We calculate a global score, \ie the cosine similarity $\langle \tau_{t_1}, \tau_{t_2} \rangle$, and an \textit{ad-hoc} per-layer metric, called \emph{subspace overlap} (simply \emph{overlap}). Let $\tau_t^{(\ell)}$ denote the weight matrix of $\tau_t$ of the layer/module $\ell$. We compute its truncated SVD and retain an orthonormal basis $V_t^{(\ell)} \in \mathbb{R}^{d \times k}$ of its top-$k$ right singular vectors. For two tasks $t_i$ and $t_j$, the overlap at $\ell$ is:
\begin{equation}\label{eq:overlap}
    \mathrm{ovl}_{\ell}(t_i,t_j) = \frac{1}{k} \bigl\| V_{t_i}^{(\ell)\top} V_{t_j}^{(\ell)} \bigr\|_F^2,
\end{equation}
with $\|\cdot \|_F$ being the Frobenius norm. In other words, $\mathrm{ovl}$ represents the mean squared cosine of the principal angles between the two rank-$k$ update subspaces, ranging from $0$ (orthogonal updates) to $1$ (coincident subspaces). The metric applies to any pair of models finetuned from a shared base model, following the same setup as the transferability analysis.




\section{Experimental Evaluation}
\subsection{Training and Optimization Settings} \label{sec:exp_setup}
We use Qwen3.5-4B~\cite{qwen3.5} as the baseline model, maintaining the same training environment ~\cite[\texttt{transformers 5.7.0},][]{wolf-etal-2020-transformers}, \cite[\texttt{TRL 1.3.0},][]{vonwerra2020trl} and hyperparameters across the specialists. Specifically, we set the learning rate at $1e-5$, 8-bit AdamW as optimizer and the batch size to $2$, with a gradient accumulation of $8$.
For the evaluation, we use \texttt{clembench 3.7.2}, with temperature set to $0$, setting \texttt{enable\_thinking} to \texttt{False}. For evaluating on single roles of two-player games, we let the corresponding specialist play against a Qwen3.5-27B model taking the other role. For the integer optimization, we use the \texttt{milp} function in the \texttt{scipy} library~\cite{scipy}.

\subsection{Transferability}
We report the resulting taxonomies in \cref{fig:qs_taxonomies}.

We can first note that visuospatial competence tasks prevalently transfer better in all budget regimes. Additionally, at $\gamma = |\mathcal{T}| = 15$ (representing the best transfer performance), the visuospatial tasks are self-sufficient, \ie they obtain their best performance without transfer (indicated by the looping edges), while verbal tasks benefit more, in general, from visuospatial task transfer (indicated by edges connecting visuospatial to verbal vertices). Taken together, these considerations suggest that the visuospatial family presents more generic challenges which help models develop versatile skills to play a broader range of games. In particular, \texttt{adventuregame} represents a good baseline specialist, due to the planning capabilities learned by the model to solve the task (\ie navigating rooms and interacting with objects). 

\subsection{Robustness of the Identified Taxonomies}
\begin{figure}[t]
    \centering
    \includegraphics[width=\linewidth]{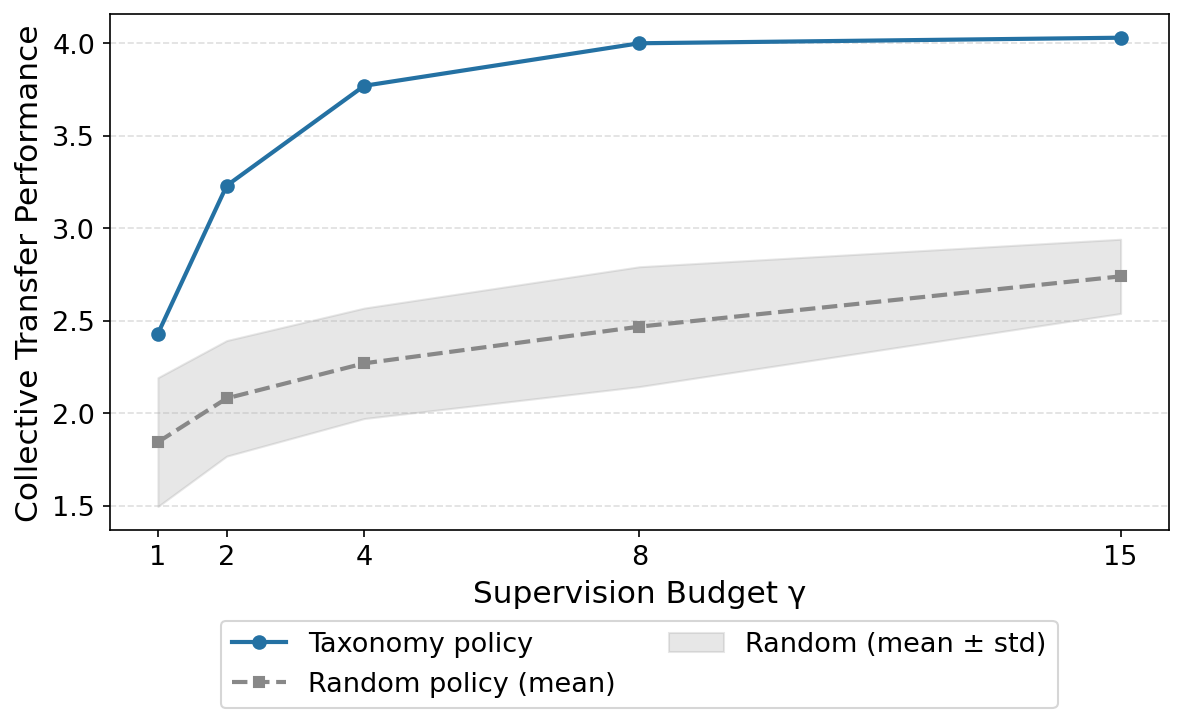}
    \caption{Comparison of the collective transfer performance against random transfer sets.}
    \label{fig:taxonomy_robustness}
\end{figure}

In order to assess whether the identified taxonomies contain meaningful information, we have compared the obtained collective transfer performance with that of $1000$ taxonomies built by randomly sampling the source tasks and the connecting edges, while maintaining the supervision budget constraint. As can be seen in \cref{fig:taxonomy_robustness}, the found transfer sets are indeed outperforming random ones by a significant margin. 

\subsection{Weight Analysis}\label{sec:weights}


The taxonomies above measure transfer by playing the games. We ask whether the same relationships can be recovered by comparing the specialists' updates directly.
We first assess if two tasks/roles related to the same game have more similarities than tasks belonging to different games. We measure a \emph{same-game overlap score} by computing the overlap metric as in \cref{eq:overlap} and averaging it over every layer-module and every pair of \texttt{P1}/\texttt{P2} tasks. Similarly, we compute a \emph{cross-game overlap score} by averaging over pairs of different game tasks. Each of the model's decoder layers contains an attention block and an MLP block; we analyse only the MLP, taking its three projection matrices (\emph{gate}, \emph{up} and \emph{down}) of the SwiGLU feed-forward block~\cite{shazeer2020glu}\footnote{The projection names follow the Hugging Face \texttt{transformers} implementation.} for the $96$ layer-module locations in total. For each location, we compare the two updated subspaces with \cref{eq:overlap}, setting $k = 5$, and average the results. We recorded a same-game overlap of $0.406$ against a cross-game overlap of $0.239$. We additionally attested that the same-game overlap beats the cross-game one at each of the $96$ layer--module locations. These results suggest, as expected, that the task vectors of two roles of the same game are far more aligned than those of two different games. Additionally, the two scores sit two orders of magnitude above the chance overlap baseline of $0.0015$ ($\approx 0.0020$ for the gate and up projections, $0.0005$ for down), \ie the expected overlap of two randomly sampled subspaces.


\begin{table}[t]
    \centering\scriptsize
    \setlength{\tabcolsep}{3pt}
    \begin{tabular}{llrrrr}
        \toprule
        task $A$ & task $B$ & $\langle\tau_A, \tau_B\rangle$ & $\text{ovl}(\tau_A, \tau_B)$ & $qs(A, B)$ & $qs(B, A)$ \\
        \midrule
        \srg{\imagegame{1}}    & \srg{\matchitascii{1}}  & $12.9$ & $44.7$ & $+10.0$ & $+6.9$ \\
        \srg{\imagegame{1}}    & \srg{\referencegame{1}} & $2.3$  & $16.5$ & $+20.0$ & $-12.9$ \\
        \srg{\imagegame{1}}    & \srg{\referencegame{2}} & $2.0$  & $12.7$ & $+0.0$  & $+3.7$ \\
        \srg{\matchitascii{1}} & \srg{\referencegame{2}} & $2.2$  & $13.3$ & $+5.0$  & $-15.0$ \\
        \skg{\taboo{1}}        & \skg{\taboo{2}}         & $7.4$  & $35.7$ & $-13.3$ & $-12.5$ \\
        \bottomrule
    \end{tabular}
    \caption{Task-vector comparison computed as cosine ($\langle \cdot, \cdot \rangle$) and top-$5$ subspace overlap ($\text{ovl}$) against $qs$ transfer in both directions. Rows are restricted to targets where the share of completed episodes is unchanged from \textit{Base}, so the quality-score gain is unambiguous. All values are reported $\times 10^2$. High similarity coincides with mutual harm (row 1) and with mutual help (row 2), so it does not determine transfer. Strong asymmetrical transfers (rows 3--5) imply low similarity.}
\label{tab:decoupling}
\end{table}

In \cref{tab:decoupling}, we compare cosine and overlap values with the $qs$ transfer scores, for some representative task pairs. While the most similar pair -- \texttt{imagegame\_P1}/\texttt{P2} -- displays good transferability, this is not always the case. Indeed,  \texttt{taboo\_P1}/\texttt{P2}, has an overall good similarity but does not transfer in any directions, suggesting that transferability is unrelated to similarity in task vectors. Conversely, the other pairs of games show asymmetric transferability, in one direction only, and score, as a consequence, lower similarities. Part of these results were to be expected: the symmetrical nature of the similarity can only partially capture transferability patterns, which are, by nature, asymmetric.

\section{Conclusions}
In this work, we have conducted an analysis on the transferability between dialogue games, adapting the methodology originally proposed by~\citet{zamir2018taskonomy}. Our experiments showed that visuospatial games display superior generalization performance compared to verbal ones, notably also to games belonging to different families. Additionally, we applied the \emph{task-vector} approach by~\citet{ilharco2022editing} to study the relationship between transferability and weight updates induced by finetuning on dialogue games, observing that similarity-based metrics are insufficient for recovering transferability from the weight space.

\section{Limitations}
The nature of this work is mainly exploratory. To confirm our results, we plan to extend our analysis both increasing the number of experiments and testing other LLMs. Additionally, we have limited our analysis to games provided by \texttt{clembench}, which are mostly collaborative. It is relevant to extend the analysis to other games, especially competitive ones. The limitation highlighted by weight-space analysis suggest going beyond symmetrical operators and explore non-symmetrical ones, that induce total order relations between games, attempting to define approaches that predict transferability without requiring evaluation on target games. Furthermore, our work focuses only on positive transfers. Expanding our taxonomy experiments to \emph{negative} knowledge transfer relationships, \ie transfers where finetuning on a certain task causes performance degradation on another one, would enable understanding how games affect each other unfavorably. Finally, it would be interesting to investigate whether the identified relationships can be exploited to build more effective curriculum-learning regimes for dialogue game training.

\bibliography{custom}

\end{document}